\documentclass[10pt,twocolumn,twoside,letterpaper]{IEEEtran}

\usepackage{geometry}
\makeatletter
\def\ps@IEEEtitlepagestyle{
  \def\@oddfoot{\mycopyrightnotice}
  \def\@evenfoot{}
}
\def\mycopyrightnotice{
  {\footnotesize
  \begin{minipage}{\textwidth}
  \centering
  978-1-7281-7693-2/20/\$31.00 \copyright2020 IEEE
  \end{minipage}
  }
}

\usepackage{cite}
\usepackage{amsmath,amssymb,amsfonts}
\usepackage{algorithmic}
\usepackage{graphicx}
\usepackage{textcomp}
\usepackage{subfigure}
\usepackage{xcolor}

\usepackage{array}
\usepackage{nopageno}
\usepackage{url}

\begin{document}

\title{SPECT Angle Interpolation Based on Deep Learning Methodologies}

\author{Charalambos Chrysostomou,
        Loizos Koutsantonis,
        Christos Lemesios
        and Costas N. Papanicolas,
\thanks{C. Chrysostomou, L. Koutsantonis, C. Lemesios and C.N. Papanicolas are with the computation-based Science and Technology Research Center, The Cyprus Institute, 20 Konstantinou Kavafi Street, 2121, Aglantzia, Nicosia, Cyprus}

}

\maketitle
\let\thefootnote\relax\footnote{Manuscript received December 20, 2020}

\begin{abstract}
A novel method for SPECT angle interpolation based on deep learning methodologies is presented. Projection data from software phantoms were used to train the proposed model. For evaluation of the efficacy of the method, phantoms based on Shepp Logan, with various noise levels added were used, and the resulting interpolated sinograms are reconstructed using Ordered Subset Expectation Maximization (OSEM) and compared to the reconstructions of the original sinograms. The proposed method can quadruple the projections, and denoise the original sinogram, in the same process. As the results show, the proposed model significantly improves the reconstruction accuracy. Finally, to demonstrate the efficacy and capability of the proposed method results from real-world DAT-SPECT sinograms are presented. 

\end{abstract}


%
\IEEEpeerreviewmaketitle

\section{Introduction}

Emission tomography is leading the medical imaging field, with Positron Emission Tomography (PET) \cite{cherry2001fundamentals, vaquero2015positron} and Single Photon Emission Computerized Tomography (SPECT) \cite{wernick2004emission, madsen2007recent, mariani2010review} as the primary techniques by detecting the trace concentrations of radioactivity. In all emission tomography methods, and more particularly in SPECT, the quality of the reconstruction is restricted by the background radiation, rate of absorption and re-scattering effects \cite{niu2011effects, ritt2011absolute, frey1994modeling}. One essential solution is to increase the radiopharmaceuticals doses given to the patients. However, the use of these radiopharmaceuticals can have unfavourable consequences on the health of the patients, and it is sought to conserve the doses to a minimum and keep the quality of the image reconstruction. Unfortunately, by reducing the radiopharmaceuticals dose on the patients, the quality of the image reconstructions is limited.  In order to address the constraints as mentioned earlier, a novel method has been proposed and developed to perform SPECT angle interpolation based on deep learning methodologies.

The paper is organised as follows: Section \ref{sec:data} presents the data generated and used for training the proposed model, Section \ref{sec:cnn}, presents the proposed model. Section \ref{sec:results}, presents the results and discussions and finally Section \ref{sec:conclussion} is conclusions.

\section{Methods and Materials}
\subsection{Training Data}
\label{sec:data}

\begin{figure}[htp]
\centering
        \includegraphics[width=9cm]{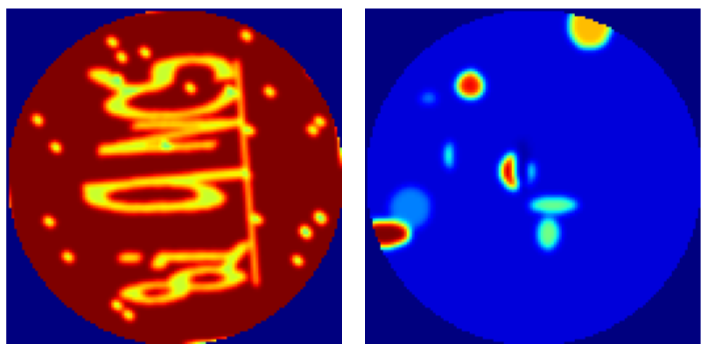}
\caption{Sample software phantoms randomly generated in order to train the proposed model}

\label{fig:random}

\end{figure}

In order to adequately train the proposed model, 200,000 software phantoms were randomly generated of $128 \times 128$ pixels. Moreover, sets of vectorised projections (sinograms) were generated from the software phantoms by simulating 32 and 128 projections, evenly spaced in 360 degrees. The generated 32 projections obtained from the phantom images were further randomised with a Poisson probability distribution to provide the noisy sets of projections. Samples of the randomly generated phantoms are shown in figure \ref{fig:random}.

\subsection{Proposed Model}
\label{sec:cnn}

Deep learning methodologies have been successfully applied in image analysis, and classification \cite{chrysostomou2018reconstruction, chrysostomou2019spect}. The proposed novel methodology is based on convolutional neural network architecture and more specifically, a U-Net architucture\cite{ronneberger2015u}. Initially, the U-Net architecture was developed for biomedical image segmentation and be very powerful for tasks like creating segmentation masks and for image processing and generation such as super-resolution or colourisation. We propose a method that can quadruple the projections and denoise the original sinogram in the same process.

\begin{figure*}[htp]
\centering
        \includegraphics[width=17cm]{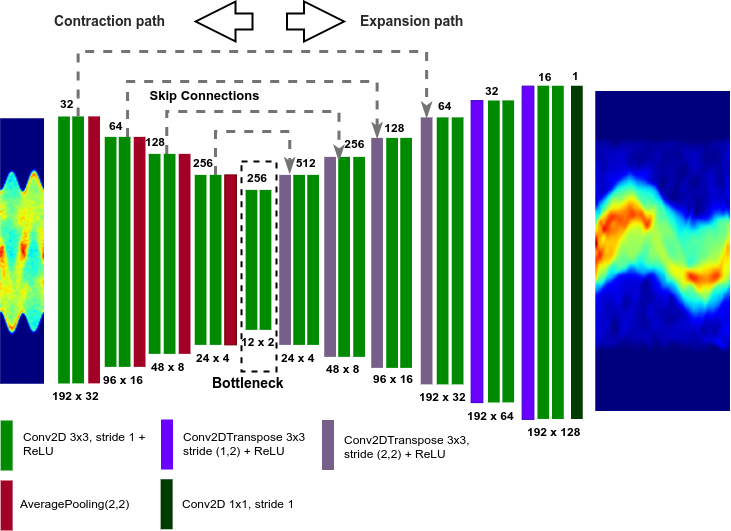}
\caption{Proposed model based on U-Net architecture for SPECT Angle interpolation.}
\label{fig:proposed_model}
\end{figure*}

The proposed method is based on standard U-Net architecture. U-Net architecture is based on fully convolutional neural networks (FCN) \cite{long2015fully} and was initially developed for biomedical image segmentation \cite{ronneberger2015u}. The advantage of using a U-Net architecture is that it can be successfully trained and operate with less training samples and generate more precise results. 

In a U-Net, architectures based on FCN, two key features are observed: (1) U-net architectures are symmetric, and (2) U-Net architectures have skip connections between the down-sampling and up-sampling layers by applying a concatenation operator. The purpose of the skip connections is to carry local information from the down-sampling stage of the model and combine it with the global information in the up-sampling stage, in order to improve the reconstruction accuracy. Figure \ref{fig:proposed_model} shows the proposed model. The proposed architecture is separated into three parts, the contracting or down-sampling path, the bottleneck and the expanding or up-sampling path. 

The contracting or down-sampling path consists of four blocks, of two convolutional layers \cite{goodfellow2016deep} each, with 3x3 kernel and rectified linear activation function (ReLU) \cite{agarap2018deep}, followed by an 2x2 average pooling layer \cite{goodfellow2016deep}. For each block, the number of kernels increases, beginning with 32 kernels for the first block, 64, 128 and 256 for the second, third and fourth blocks respectively. The main objective of the contracting path is to obtain the core context of the input sinogram in order to perform the denoising. The common contextual information of the sinograms will be carried to the upsampling path using skip connections. The bottleneck part of the proposed architecture consists of two convolutional layers, with 3x3 kernel and ReLU as the activation function.

The expanding or up-sampling path consists of seven blocks.  The first six blocks consist of a transposed convolutional layer and two convolutional layers with 3x3 kernels and ReLU as the activation function. The last block is a single convolutional layer of 1x1 kernel. The first four blocks of the expanding path are to allow accurate translation of the core context of the sinograms and to connect the contextual information carried using skip connections from the contracting path. The last three blocks are used to combine the core context and contextual information accurately into a denoised sinogram with quadruple the projections.

The advantage of using the proposed model is the ability to connect contextual information from the up-sampling path with the location information from the down-sampling path, which is essential to perform SPECT angle interpolation. An additional benefit of the proposed model is that no dense layers are used, which allows variable sizes of sinograms. The only parameters that need to be optimised are the kernel in the convolutional layers which are independent of input size.

\section{Results and Discussions}
\label{sec:results}

\begin{figure}[htp]
\centering
        \includegraphics[width=8cm]{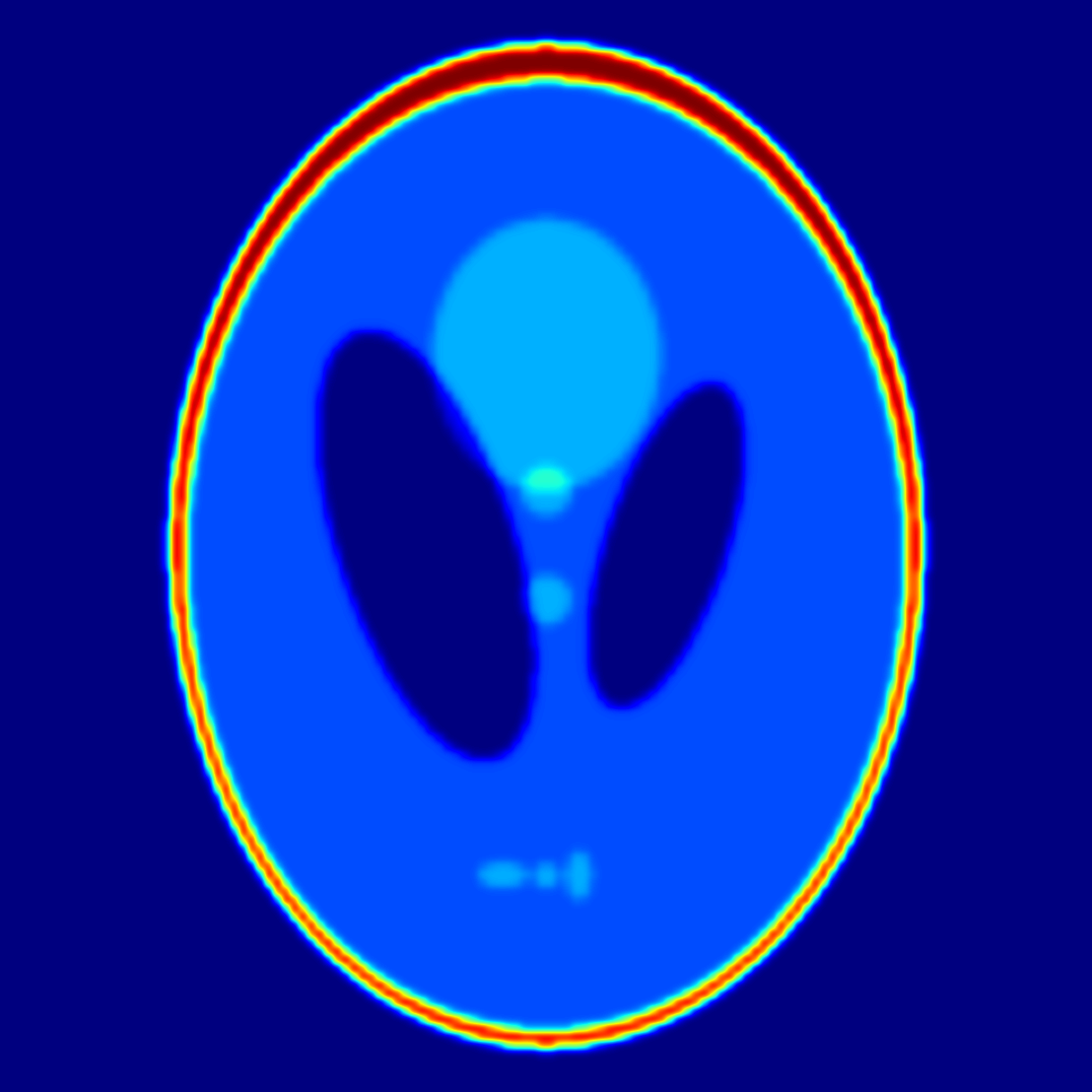}
\caption{Shepp Logan Phantom used to evaluate and demonstrate the capabilities of the proposed method}
\label{fig:original_phantom_sinogram}
\end{figure}
\begin{figure*}[htp]
\centering
\includegraphics[width=17cm]{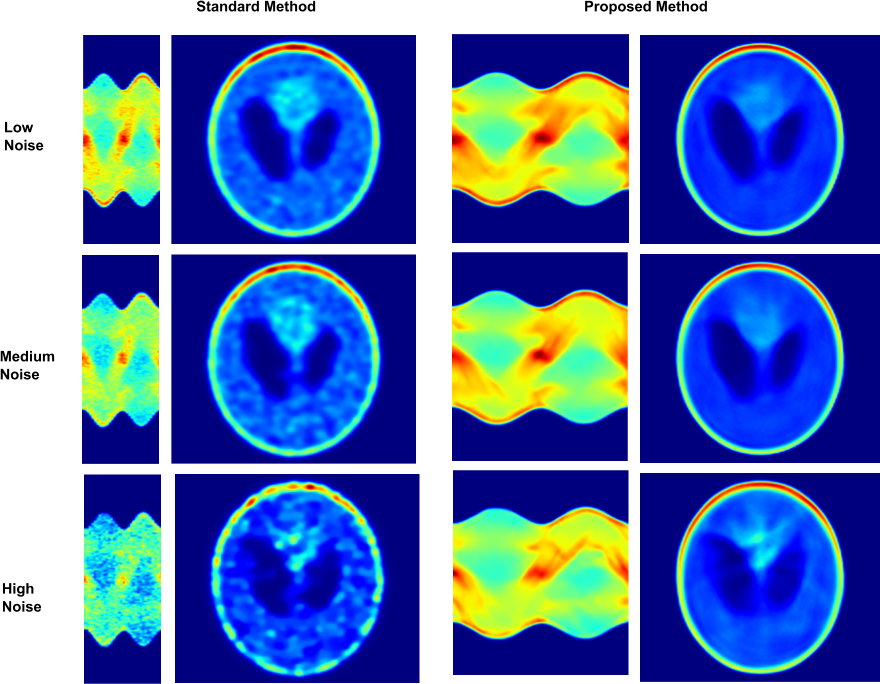}
  \caption{Results of the sinogram reconstructions by using OSEM for the proposed method versus the standard method}
\label{fig:results}
\end{figure*}
\begin{figure*}[htp]
\centering
\includegraphics[width=17cm]{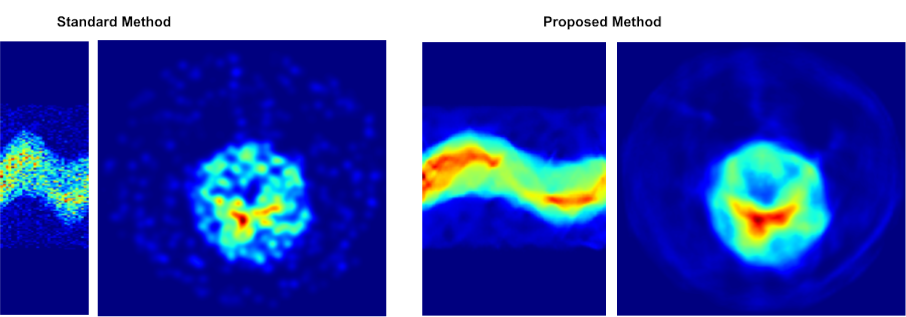}
  \caption{Results of the DAT-SPECT sinogram reconstructions by using OSEM for the proposed method versus the standard method}
\label{fig:real_results}
\end{figure*}

The efficiency of the proposed method and comparison with existing methods was performed using, the Mean Absolute Percentage Error (MAPE), Mean Square Error (MSE), Structural Similarity (SSIM) Index \cite{wang2004image}, and the Peak signal-to-noise ratio (PSNR) were used. The Shepp Logan phantom \cite{shepp1974fourier}, as shown in Fig. \ref{fig:original_phantom_sinogram} was used to demonstrate and evaluate the capabilities of the proposed method. Table \ref{denoise_results_table} shows the results of the denoised method for the three different levels of noise compared to the original, noise-free 128 projections sinogram. Table \ref{osem_results_table} and Figure \ref{fig:results} shows the reconstruction results based on the standard OSEM method versus the proposed method. As the results show, the proposed method significantly denoises the sinograms and improves the reconstructions, in comparison with the original 32 projections sinogram. Finally, in order to test the efficacy of the proposed model, we used real-world DAT-SPECT sinograms and produced reconstructions based on the standard OSEM and proposed method. The results are presented in Fig. \ref{fig:real_results}.  As the results show, the proposed model is capable of been applied and improve real-world problems, where the collection of training data is limited or challenging, by using computationally inexpensive software phantoms.

\begin{table}[ht]
\renewcommand{\arraystretch}{1.5}
\centering
\caption{Sinogram DeNoising Results}
\label{denoise_results_table}

\begin{tabular}{lllll}
\hline

\textbf{Noise} &  \textbf{MAPE} &\textbf{MSE} & \textbf{SSIM} & \textbf{PSNR} \\
\hline

Low            & 2.64\%      & 0.0006       & 0.977         & 32.51         \\
Medium         & 4.13\%      & 0.0009       & 0.972         & 30.64         \\
High           & 6.34\%       & 0.0018       & 0.948         & 27.51        \\
\hline

\end{tabular}
\end{table}

\begin{table}[ht]
\renewcommand{\arraystretch}{1.5}
\centering
\caption{Reconstruction Results based on OSEM}
\label{osem_results_table}

\begin{tabular}{lllllll}
\hline
& \multicolumn{3}{c}{\textbf{Standard Method}} & \multicolumn{3}{c}{\textbf{Proposed Method}} \\
\hline
\textbf{Noise Level} & \textbf{MSE}    & \textbf{SSIM}    & \textbf{PSNR}   & \textbf{MSE} & \textbf{SSIM} & \textbf{PSNR} \\
Low                  & 0.0071          & 0.84             & 21.49           & 0.0039       & 0.90          & 24.07         \\
Medium               & 0.0076          & 0.82             & 21.20           & 0.0036       & 0.90          & 24.46         \\
High                 & 0.0093          & 0.76             & 20.33           & 0.0051       & 0.86          & 22.94  \\
\hline
\end{tabular}

\end{table}

\section{Conclusions}
\label{sec:conclussion}

In this study, we demonstrate the capability of the convolutional neural networks to perform SPECT angle interpolation in SPECT imaging. The proposed method was trained using software phantoms, and the capabilities of the system were tested using software-generated sinograms as well as real-world DAT-SPECT sinograms. As the results show, the proposed method significantly improves the results and outperforms the standard OSEM method. Although the phantom and the results presented in this paper is appropriate for illustrating the capabilities of the proposed method, further experimentation is needed for evaluating the potential application of the method in clinical studies.

\bibliography{ref} 
\bibliographystyle{IEEEtran}

\end{document}